\documentclass[twoside,11pt]{article}

\usepackage[abbrvbib, preprint]{jmlr2e}

\usepackage{jmlr2e}
\usepackage{hyperref}
\usepackage{natbib}
\usepackage{graphicx}
\usepackage{physics}
\usepackage{xcolor}
\usepackage{amsmath}
\usepackage{amssymb}
\usepackage{tikz}
\usepackage{caption}
\usepackage{subcaption}
\usepackage{comment}
\usepackage[normalem]{ulem}

\hypersetup{
    citecolor=black,
    colorlinks=true,
    linkcolor=black,
    filecolor=blue,      
    urlcolor=blue,
}

\ShortHeadings{Loss Landscape Engineering via Data Regulation on PINNs}{Vignesh, Stanislas and Debasmita}
\firstpageno{1}

\begin{document}

\title{Loss Landscape Engineering via Data Regulation on PINNs}

\author{\name Vignesh Gopakumar \email  vignesh.gopakumar@ukaea.uk \\
      \addr UK Atomic Energy Authority\\
      Culham Science Centre \\
      Abingdon, Oxfordshire, OX14 3DB, United Kingdom
      \AND
      \name Stanislas Pamela \email  stanislas.pamela@ukaea.uk \\
      \addr UK Atomic Energy Authority\\
      Culham Science Centre \\
      Abingdon, Oxfordshire, OX14 3DB, United Kingdom
      \AND
      \name Debasmita Samaddar \email  dsamaddar@alaska.edu \\
      \addr UK Atomic Energy Authority\\
      Culham Science Centre \\
      Abingdon, Oxfordshire, OX14 3DB, United Kingdom}

\maketitle

\begin{abstract}%   <- trailing '%' for backward compatibility of .sty file
Physics-Informed Neural Networks have shown unique utility in parameterising the solution of a well-defined partial differential equation using automatic differentiation and residual losses. Though they provide theoretical guarantees of convergence, in practice the required training regimes tend to be exacting and demanding. Through the course of this paper, we take a deep dive into understanding the loss landscapes associated with a PINN and how that offers some insight as to why PINNs are fundamentally hard to optimise for. We demonstrate how PINNs can be forced to converge better towards the solution, by way of feeding in sparse or coarse data as a regulator. The data regulates and morphs the topology of the loss landscape associated with the PINN to make it easily traversable for the minimiser. Data regulation of PINNs helps ease the optimisation required for convergence by invoking a hybrid unsupervised-supervised training approach, where the labelled data pushes the network towards the vicinity of the solution, and the unlabelled regime fine-tunes it to the solution
\end{abstract}

\begin{keywords}
  Physics-Informed Neural Networks, Loss Landscape, Sparse regularization, Partial Differential Equations
\end{keywords}

\section{Physics Informed Neural Networks - Introduction}

Physics Informed Neural Networks (PINNs) are commonly networks that are deployed to map the spatio-temporal coordinates to the field variables associated with a well defined Partial Differential Equation (PDE) or a family of PDEs. The networks are trained in an unsupervised manner, constrained with modified loss functions that embed the dynamics prescribed by the PDE in the form of residuals, the initial distribution of the field variables and the associated boundary conditions \cite{RAISSI2019}. \\

Consider a nonlinear Partial Differential Equation of the the general form:

\begin{equation}
    \Gamma(u, t) + \Lambda(u, X) = 0,\quad X \ \epsilon\ \Omega,\ t\ \epsilon\ [0, T]
    \label{eq: gen_pde}
\end{equation}
\newline
where $t$ represents the temporal coordinate time confined by a domain $[0,T]$, $X$ represents the spatial coordinates which belongs to the space $\Omega$, a subset of $\mathbb{R}^D$ ($D$ represents the number of spatial dimensions), $u(X,t)$ refers to the field variable(s) of interest that are being modelled by the PDE. $\Gamma$ is a function that represents the amalgamation of all the partial derivatives of the field variable with respect to time, while $\Lambda $ is a function that accounts for all other terms within the PDE, including spatial derivatives or eventual non-linear terms. \\

To further elucidate the formulation given in equation \ref{eq: gen_pde}, consider the two dimensional Wave equation, $\pdv[2]{u}{t} = c\big(\pdv[2]{u}{x} + \pdv[2]{u}{y}\big)$, the equation can be compared to the formulation as:  $\Gamma(u,t) = \pdv[2]{u}{t}$, while $\Lambda(u, X) = - c\big(\pdv[2]{u}{x} + \pdv[2]{u}{y}\big)$. \\

To obtain well-defined solutions of a system of PDEs, the equations are coupled with an initial condition that describes the initial distribution of the field variables as well as the boundary conditions: 

\begin{align}
    \text{Initial Condition:  } u(X_i, 0) = f(X_i),\  X_i\ \epsilon\ \Omega\label{eq: ic} \\
    \text{Boundary Condition:  } u(X_b, t) = g(X_b, t),\ X_b\ \epsilon\ \partial\Omega,\  t\ \epsilon\ [0, T] \label{eq: bc}
\end{align}
where, $f$ and $g$ are arbitrary functions, and where $\partial\Omega$ is the boundary of the domain $\Omega$. \\

\begin{figure}[h!]
    \centering
    \includegraphics[scale=0.35]{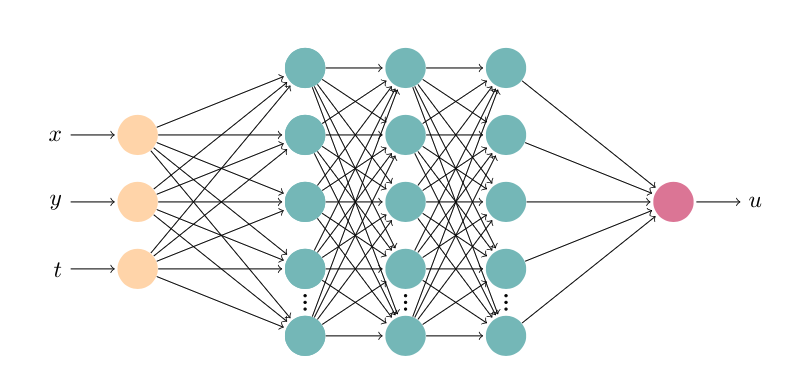}
    \caption{Layout of a Physics Informed Neural Network mapping from the spatio-temporal input space to the output field space.}
    \label{fig:PINN_Arch}
\end{figure}
In order to solve the well-defined PDE as given by equations \ref{eq: gen_pde}, \ref{eq: ic} and \ref{eq: bc} using a PINN approach, a fully-connected neural network with nonlinear activation functions as given in figure \ref{fig:PINN_Arch} is constructed. The network takes in three inputs, the spatio-temporal coordinates $x,y,t$ and outputs the field variable $u$. The network is trained in an unsupervised manner by minimising the residual error associated with the PDE, initial and boundary conditions by way of a modified loss function. This differs from supervised learning scenarios, where the loss function is governed by the reconstruction error, defined by the deviation of the neural network output from the true output as given in the training dataset. However, for a PINN we do not employ a labelled dataset but instead measure the loss by the deviation of the neural network output from the governing equations. The loss function of a PINN can be given as:

\begin{equation}
    \text{[Training Loss] = [Domain Loss] + [Initial Loss] + [Boundary Loss]}.
    \label{eq: training_loss}
\end{equation}
Substituting equations \ref{eq: gen_pde}, \ref{eq: bc} and \ref{eq: ic} into equation \ref{eq: training_loss}, we obtain:
\begin{equation}
  \begin{split}
        \text{[Training Loss]} =  \Gamma(\tilde{u}, t) + \Lambda(\tilde{u}, X_d) + \\
        \tilde{u}(X_i, 0) - f(X) + \\
        \tilde{u}(X_b, t) - g(Y, t)
    \end{split}  
    \label{eq: training_loss_expanded}
\end{equation}
where $\tilde{u}$ is the neural network output and $X_i, X_b, X_d$ coupled with the appropriate $t$ are the neural network inputs. \\ 

Equation \ref{eq: training_loss_expanded} estimates how well the neural network satisfies the PDE and the its associated constraints. In order to estimate the local partial derivatives of the field variables in space and time as found in equation \ref{eq: training_loss_expanded}, Automatic Differentiation tools are employed \cite{BAYDIN2018}. 

\subsection{PINNs over Numerical Schemes}
Traditional numerical methods (finite difference, finite element methods) build solutions to PDEs bounded to a mesh, and inherently provide discretised solutions. PINNs on the other hand are bounded by the domain of interest that the input spatio-temporal coordinates $(X,t)$ span, hence rendering PINN solutions to be mesh-free and continuous. In addition, traditional numerical methods require precise implementation of a numerical solver, which can vary significantly depending on the PDEs and coordinates systems of the problem, while PINNs have a standard setup which remains mostly unchanged except for the formulation of the loss function as we move across various cases and physics-scenarios. PINNs are also invariant to change in the coordinate system as these are automatically accounted for within the loss function. Finally, traditional numerical schemes often pose significant difficulty in being parallelised across devices, whereas a PINN, being reliant on a neural network architecture, can easily be deployed across an ensemble of GPUs and convergence to solution can be accelerated. \\

\subsection{Visualising the Loss Landscape}
Empirical understandings of the efficacy of the trained model with respect to the minima associated with the objective function, can be extensively visualised by mapping the loss landscape around the minimiser to which the PINN. We perform this by projecting the weights of the trained network along two orthogonal directions (\cite{GOLDSTEIN2016}, Lemma 5). The model being updated with the weight projections are then evaluated within the criteria of the loss function across this 2D projection space and mapped. For an extensive description of the visualisation method refer \cite{LI2018}.  \\

A network initialised with parameters $\theta$, are trained to a model state characterised by $\theta^*$. Two orthogonal vectors are chosen in this parameter space $\delta_1$ and $\delta_2$. The weight parameters are projected as: 

\begin{equation}
    \theta^*_{proj} = \theta^* + \alpha\delta_1 + \beta\delta_2
\end{equation} 

In order to deal with the scale invariance properties of neural networks, we deploy a filter-wise normalisation, ensuring the directional weights $\delta_1$ and $\delta_2$ has the same norm as that of the PINN weights $\theta^*$ \cite{LI2018}.

\subsection{Complexities and Errors}
Neural Networks with sufficient neurons can simultaneously and uniformly approximate any function and its partial derivatives \cite{LU2019}. However, when we fixate on neural network of a specific architecture, we constrain and limit the space of nonlinear mappings that can be performed by the network (approximation error). The training dataset (i.e. the input points gathered across the domain) contributes to the second limiting factor. The solution to which the network converges to is governed by these points, and often might disproportionately represent the nonlinear mappings associated with the PDE (generalisation error). Hence the global minimum associated with the training dataset might not adequately represent the solution of the well-defined PDE which occupies a unique position in the loss landscape. Training the neural network towards the PDE solution often typically relies on stochastic gradient descent methods which engage in non-convex optimisation, often leading to the network getting stuck in slightly favourable local minima \cite{BLUM1992} (optimisation error). Bottou \textit{et al.} demonstrated that in a small-scale learning problem there exists a trade-off between approximation and estimation \cite{BOTTOU2008}. Lu \textit{et al.} explored this relationship within a PINN framework and showcased that a PINN solution carries with it inherent errors which can be expressed as \cite{LU2019}: 

\begin{equation}
    \epsilon = \epsilon_{approx} + \epsilon_{general} + \epsilon_{optim}
\end{equation}
\newline
A well-defined PDE often has a unique solution that is prescribed by the initial-boundary value problem setup. For a PINN to effectively converge to the solution, it requires achieving this unique point in the arbitrary loss landscape that is defined by the network architecture, training dataset and the loss function together in conjunction. Loss landscapes facilitated by complex multi-objective loss functions (as seen in equation \ref{eq: training_loss_expanded}) involving local and higher order derivatives often have complicated Pareto-optimal fronts \cite{NGATCHOU2005}. These multi-objective loss entities tend to compete against each other within a restricted Pareto front with little to no room for movement. Non-convex optimisation poses a certain computational intractability \cite{BLUM1992}, PINNs guided by the PDE constraints often get stuck in local minima found across this tumultuous loss landscape, leading to large optimisation errors. \\

Since the objective of a PINN is to solve for a well-defined PDE, it would be in our interest to ignore the generalisation error and train the model to develop a heavy bias towards the solution. This bias towards the solution ignoring the generalisation error leads to a decrease in the approximation error allowing for better convergence. A PINN can be biased towards the solution by adding simulation/experimental data as a regulator within the training regime. Simulation data points, either sparsely sampled or gathered from a coarse solver for this purpose can be employed to form a reconstruction error, moving the optimisation task from that solely involving unsupervised learning to a hybrid regime that accommodates for unsupervised and supervised learning. \\

The paper is structured as follows: In section \ref{sparse pinns}, we explore the impact of regulating the PINNs with sparsely sampled simulation data, discuss how that affects the performance as well as the topological changes made to the loss landscape. We also explore the changes that occurs as we increase the amount of sparse data made available within the training regime. In section \ref{coarse pinns}, we explore the impact data gathered from coarse simulations can have in training PINNs to solve the PDE on a non-discretised domain. Section \ref{experimental data} discusses the impact psuedo-experimental data gathered from a fixed point in the spatial domain, akin to a diagnostic tool within an experimental setup can have on fine-tuning the PINN. Finally in section \label{conclusion}, we conclude the paper by summarising the results of our experiments, highlight our contributions, engage in a discussion about the advantages and disadvantages as well as provide an outlook to future work. \\

\section{Sparse Regulated PINNs}
\label{sparse pinns}
A sparse regulated PINN is identical in architecture to a Vanilla PINN as demonstrated in \cite{RAISSI2019}, \cite{LU2019}, \cite{HENNIGH2020}, \cite{SUN2020}, with the the only modification being in the formulation of the loss function. In addition to the three PDE constraints as demonstrated in equation \ref{eq: training_loss_expanded}, we have a new objective added to it, reconstruction loss of sparse simulation data points. The loss function for a sparse regulated PINN is expressed as:  
\begin{equation}
  \begin{split}
        \text{[Sparse Training Loss]} =  \Gamma(\tilde{u}, t) + \Lambda(\tilde{u}, X_d) + \\
        \tilde{u}(X_i, 0) - f(X_d) + \\
        \tilde{u}(X_b, t) - g(X_b, t) + \\
        \tilde{u}(X_s,t) - u(X_s,t) 
    \end{split}  
    \label{eq: sparse_training_loss}
\end{equation}
where, in addition to the terms described in equation \ref{eq: training_loss_expanded}, $\tilde{u}(X_s,t)$ is the neural network output for the sparse data points $X_s$, while $u(X_s,t)$ is the actual solution representing the field variable at space $X_s$ and time $t$. \\

Randomly gathering relatively small portions of sparsely located solution data from across the domain has vast impacts on reducing the approximation error. But it turns out that it is not only the approximation error that is reduced by regulating the training with sparse data. They work to decrease the optimisation error as well. Optimisation error is decreased as the sparse data points fed in as a regulator within the training regime morph the topology of the loss landscape, making it easily traversable for the optimiser. This holds true only of the sparse data is accurate, however if it is generated from a less-refined model, we must account for coarseness of the fit by weighing the loss function. \\

\subsection{Examples}
For all experiments run on PINNs and Sparse regulated PINNs we use the same architecture as found in figure \ref{fig:PINN_Arch}. For each simulation, we deploy a ResNet based architecture with two blocks \cite{HE2015}, each block consisting of two fully connected layers with 64 neurons each. The ResNet configuration has been chosen to avoid dealing with issues of vanishing gradients as we require differentiating to higher order derivatives to satisfy the PDE criterion. A fully connected layer is deployed as an interface between the residual blocks and the output layer. We employ a Tanh activation function after each fully connected layer and block to effectively model non-linearity. Tanh functions are suited for PINNs as they are able to preserve higher order gradients \cite{RAMACHANDRAN2017}. All networks (except for Burgers') are trained for 20000 epochs with the Adam optimiser \cite{KINGMA2017} using a step scheduler for learning rate starting at 1e-3 and decreasing by a gamma factor of 0.9 every 5000 steps. A Quasi-Monte Carlo method is applied across each entity of the loss function to be able to integrate over the respective domains for discrete batches of the spatio-temporal inputs \cite{HENNIGH2020}.

\subsubsection{Burgers' Equation}
Consider the one dimensional Burgers' equation:
\begin{align*}
    \pdv{u}{t} + u\pdv{u}{x} - \pdv[3]{u}{x}, \quad x \ \epsilon\ \Omega,\ t\ \epsilon\ [0, 1] \\ 
    u(x,t=0) = -\sin(\pi x) + 1/\cosh(x) \\
\end{align*}
where, $\Omega\ \epsilon\ [-1,1]$, $\nu = 0.01/\pi$ and bounded periodically.\\

The solution for the above equation is built using a spectral solver implemented in python. Comparing the solutions across the Vanilla PINN and Sparse regulated PINNs trained with 1$\%$ sparse data (as seen in figure \ref{fig:wave_solution}), it is evidently clear how the sparse data helps achieve better performance. 

\begin{figure}[h!]
    \centering
    \includegraphics[scale=0.25]{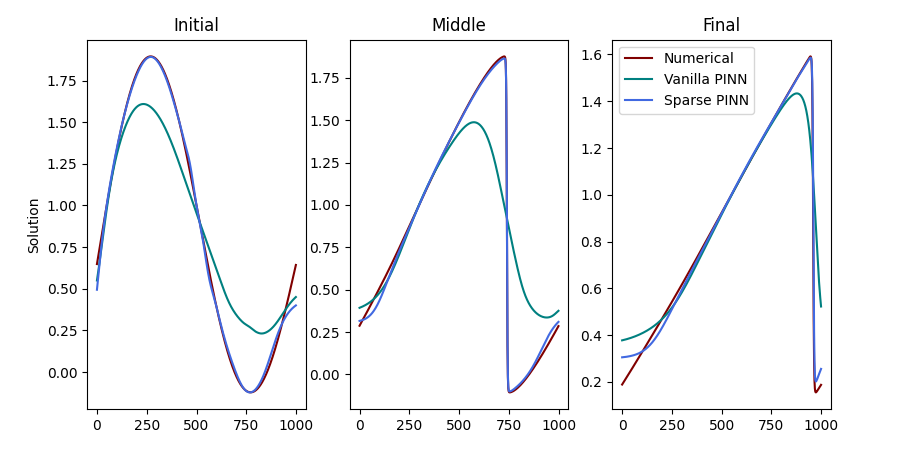}
    \caption{PINN solutions visualised and compared to the numerical solution. The plot shows the network outputs at the time = 0 (initial), time=0.5 (middle) and time=1.0 (final). The Vanilla PINN trains to a L2 Error of 0.078 while the Sparse Regulated PINN converges to 0.001. The training was terminated after 5000 epochs. }
    \label{fig:wave_solution}
\end{figure}

\begin{figure}[h!]
     \centering
     \begin{subfigure}{0.45\textwidth}
         \includegraphics[scale=0.3]{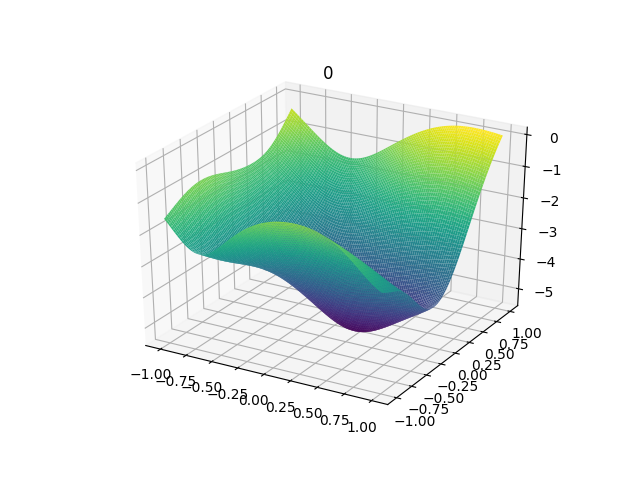}
         \caption{No Sparse Data}
         \label{sparse_0}
     \end{subfigure}
     \hfill
     \begin{subfigure}{0.45\textwidth}
         \includegraphics[scale=0.4]{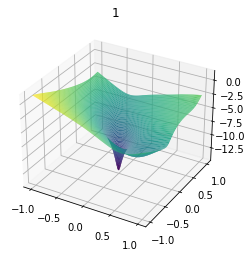}
         \caption{1\% Sparse Data}
         \label{sparse_1}
     \end{subfigure}
        \caption{Loss landscapes visualised of trained PINNs solving the 1D Burgers equation with and without sparse data representations. Figure \ref{sparse_0} shows a clear minima with a relatively bumpy loss landscape for the case of a Vanilla PINN, but in figure \ref{sparse_1} when sparse data representations are introduced the landscape becomes more defined. The X and Y axes represent projection of the trained neural network within orthogonal directions within the weight space, while the Z axis represents the logarithm of the loss value along these projections. }
        \label{fig:landscapes_burgers}
\end{figure}

The difference in performance can be further substantiated once we map the loss landscape associated with each PINN as seen in figure \ref{fig:landscapes_burgers}. The influence of the sparse data regulates and accentuates the crests and troughs of the landscape, helping the minimiser to reach a better minima. 

\newpage

\subsubsection{Wave Equation}

Consider the two dimensional wave equation:

\begin{align*}
    \pdv[2]{u}{t} - \bigg(\pdv[2]{u}{x} + \pdv[2]{u}{y}\bigg) = 0 , \quad x,y \ \epsilon\ \Omega,\ t\ \epsilon\ [0, 1]\\
    u(x,y,t=0) = \exp^{-40((x-4)^2 + y^2)} \\
    \pdv{u(x,y,t=0)}{t} = 0\\
    u(x,y,t) = 0, \quad x,y \ \epsilon\ \partial\Omega,\ t\ \epsilon\ [0, 1]
\end{align*}
where, $\Omega\ \epsilon\ [-1,1]$ \\

The solution for the above equation is built by deploying a spectral solver that uses a leapfrog method for time discretisation and a Chebyshev spectral method on tensor product grid for spatial discretisation \cite{wave_spectral}.\\

\begin{figure}[h!]
    \centering
    \includegraphics[scale=0.5]{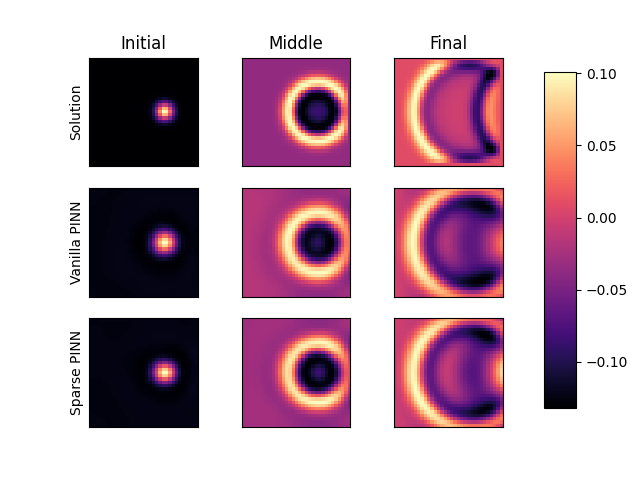}
    \caption{PINN solutions visualised and compared to the numerical solution. The plot shows the network outputs at the time = 0 (initial), time=0.5 (middle) and time=1.0 (final). The Vanilla PINN trains to a L2 Error of  $5\times10^{-4}$ while the Sparse Regulated PINN converges to $3\times10^{-4}$.}
    \label{fig:wave_solution}
\end{figure}

Initially we train a vanilla PINN, optimised by enforcing the PDE constraints alone. This is followed up by training a sparse regulated PINN which in addition to the PDE constraints accommodates 1 percent of randomly sampled solution data to the training regime. Upon evaluating the loss landscape of both the trained networks and visualising we can see that the sparse data representations introduce necessary topological changes to the landscape creating succinct features making it easier for the minimiser to traverse and reach the minima. \\

\begin{figure}[h!]
     \centering
     \begin{subfigure}{0.45\textwidth}
         \includegraphics[scale=0.3]{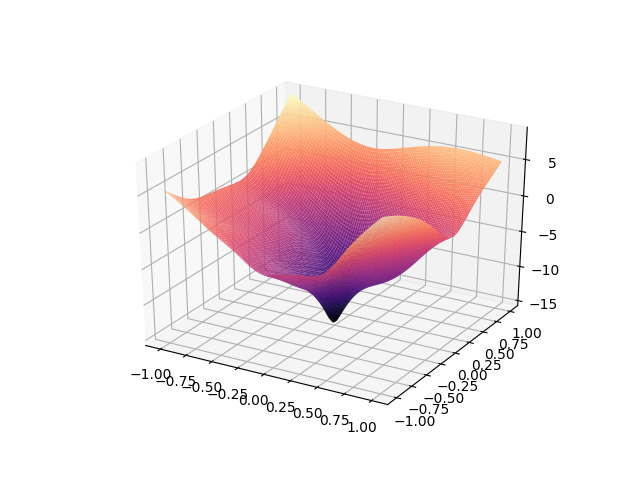}
         \caption{No Sparse Data}
         \label{sparse_0}
     \end{subfigure}
     \hfill
     \begin{subfigure}{0.45\textwidth}
         \includegraphics[scale=0.3]{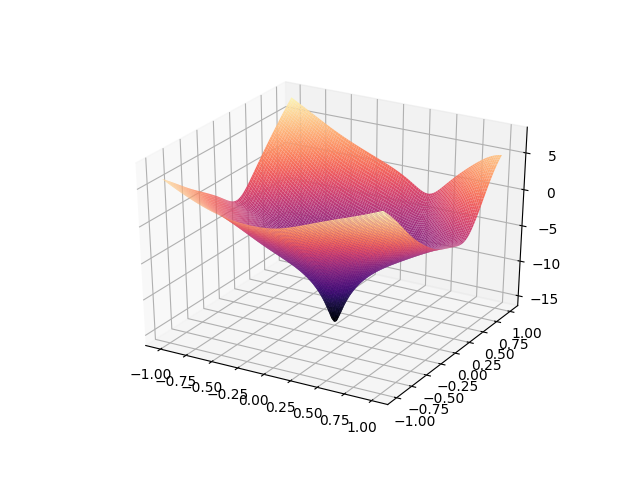}
         \caption{1\% Sparse Data}
         \label{sparse_1}
     \end{subfigure}
        \caption{Loss landscapes visualised of trained PINNs solving the 2D Wave Equation with and without sparse data representations. Figure \ref{sparse_0} shows a clear minima with a relatively bumpy loss landscape for the case of a Vanilla PINN, but in figure \ref{sparse_1} when sparse data representations are introduced the landscape becomes more defined with sharper minima. The X and Y axes represent projection of the trained neural network within orthogonal directions within the weight space, while the Z axis represents the logarithm of the loss value along these projections. }
        \label{fig:landscapes}
\end{figure}

\subsubsection{Navier-Stokes Equation}
Consider the two dimensional Navier-Stokes equation:

\begin{align*}
    \pdv{u}{t} + u\pdv{u}{x} + v\pdv{u}{y}  + \frac{1}{\rho}\pdv{p}{x} - \nu\bigg(\pdv[2]{u}{x} + \pdv[2]{u}{x}\bigg) = 0 \\
    \pdv{v}{t} + u\pdv{v}{x} + v\pdv{v}{y}  + \frac{1}{\rho}\pdv{p}{y} - \nu\bigg(\pdv[2]{v}{x} + \pdv[2]{v}{x}\bigg) = 0 \\
    \pdv[2]{p}{x} + \pdv[2]{p}{y} + \rho\bigg( \big(\pdv{u}{x}\big)^2 + 2\pdv{u}{x}\pdv{v}{y} + \big(\pdv{v}{y}\big)^2\bigg) =0  \\
    \quad x,y \ \epsilon\ \Omega,\ t\ \epsilon\ [0, 1]
\end{align*}
where, $\Omega\ \epsilon\ [0,20]$x$[0,10]$ and $ \nu = 0.04, \rho=1.0, Re = 50$\\

The equations are employed along with the necessary initial and boundary conditions to model fluid flow around a rectangular block. The numerical solution is built using an explicit finite difference solver implemented with the FTCS scheme \cite{TANNEHILL1997}. 

\begin{figure}[h!]
    \centering
    \includegraphics[scale=0.5]{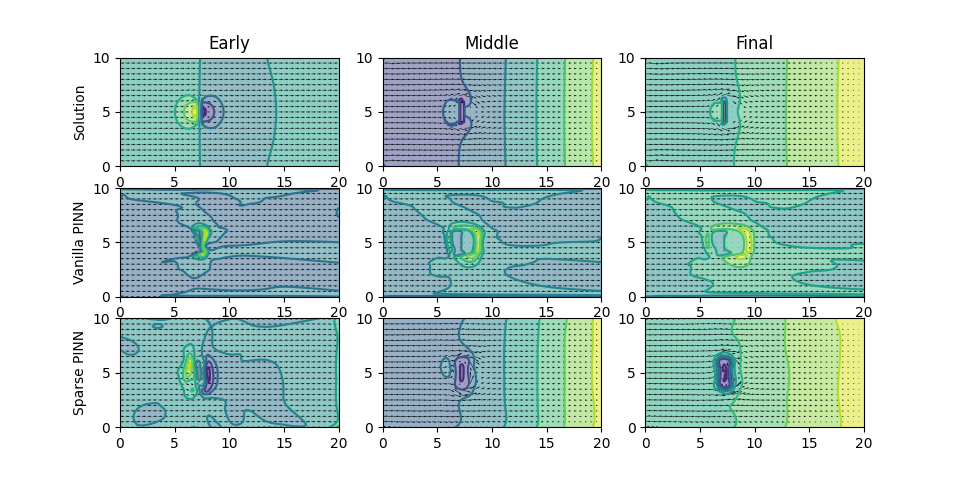}
    \caption{PINN solutions visualised and compared to the numerical solution. The plot shows the network outputs at the time = 0.01 (early), time=0.5 (middle) and time=1.0 (final). The Vanilla PINN trains to a L2 Error of 8.237 while the Sparse Regulated PINN converges to 0.2331.}
    \label{fig:ns_solution}
\end{figure}

For the given PDE setup we notice that the Vanilla PINN struggles to converge towards the solution and is only capable of identifying the mere presence of the block, whereas the sparse regulated PINN models a lot more and gets to the vicinity of the true solution. The convergence properties of both the vanilla and sparse regulated PINNs can be characterised by looking at the topology of the loss landscape. \\

\begin{figure}[h!]
     \centering
     \begin{subfigure}{0.45\textwidth}
         \includegraphics[scale=0.3]{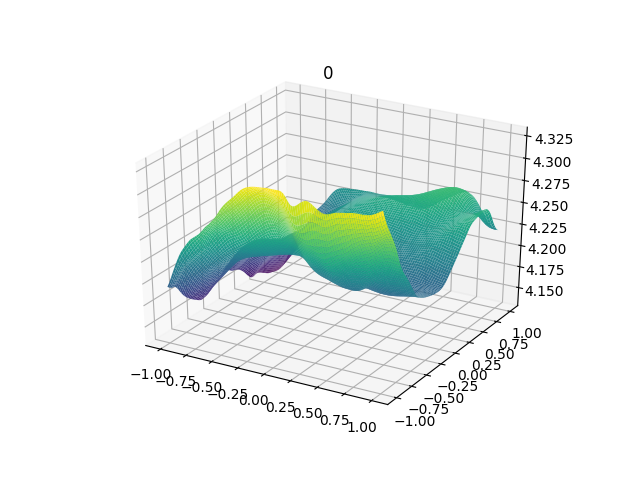}
         \caption{No Sparse Data}
         \label{sparse_0_ns}
     \end{subfigure}
     \hfill
     \begin{subfigure}{0.45\textwidth}
         \includegraphics[scale=0.3]{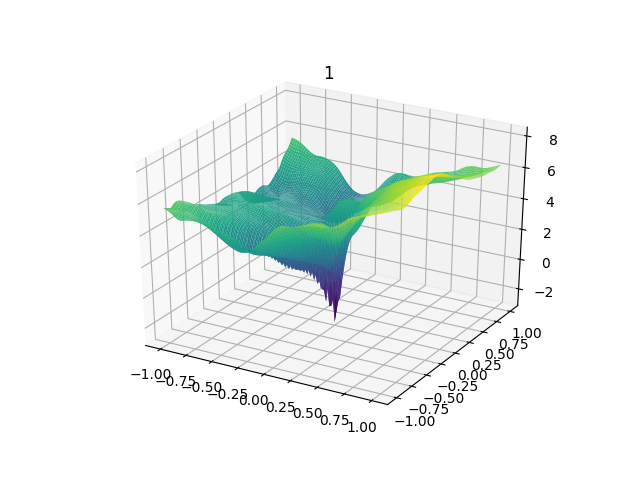}
         \caption{1\% Sparse Data}
         \label{sparse_1_ns}
     \end{subfigure}
        \caption{Loss landscapes visualised of trained PINNs solving the 2D Navier-Stokes modelling flow around block with and without sparse data representations. Figure \ref{sparse_0_ns}, there is no minima in sight and the model would require further extensive training to reach a minima. Figure \ref{sparse_1_ns} shows a clear overhaul of the loss landscape performed by the sparse simulation data. The model is currently stuck in a local minima and can reach the better minima within sight with a little training. The X and Y axes represent projection of the trained neural network within orthogonal directions within the weight space, while the Z axis represents the logarithm of the loss value along these projections. }
        \label{fig:landscapes_ns}
\end{figure}

\begin{figure}[h!]
     \centering
     \begin{subfigure}{0.45\textwidth}
         \includegraphics[scale=0.3]{images/NS_LL_0.png}
         \caption{No Sparse Data}
         \label{sparse_0_ns}
     \end{subfigure}
     \hfill
     \begin{subfigure}{0.45\textwidth}
         \includegraphics[scale=0.3]{images/NS_LL_1.png}
         \caption{1\% Sparse Data}
         \label{sparse_1_ns}
     \end{subfigure}
     \begin{subfigure}{0.45\textwidth}
         \includegraphics[scale=0.3]{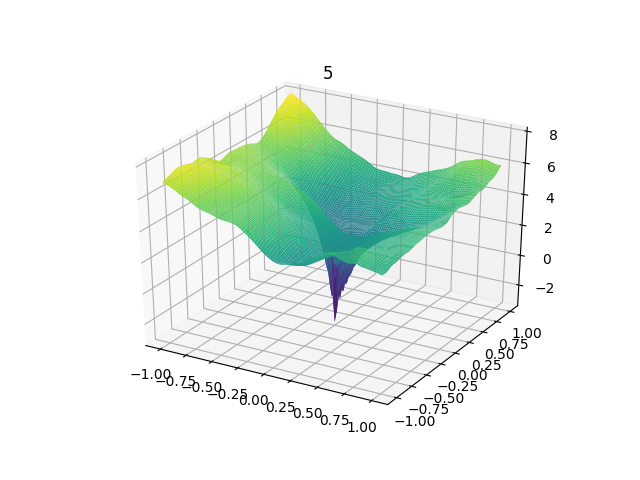}
         \caption{5\% Sparse Data}
         \label{sparse_5_ns}
     \end{subfigure}
     \begin{subfigure}{0.45\textwidth}
         \includegraphics[scale=0.3]{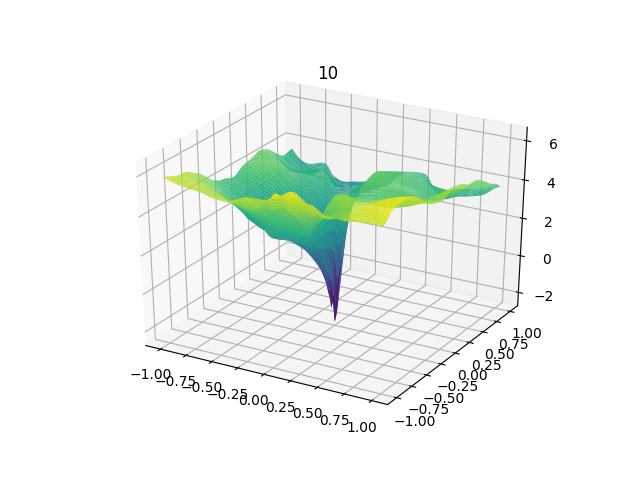}
         \caption{10\% Sparse Data}
         \label{sparse_10_ns}
     \end{subfigure}
        \caption{Loss landscape visualistions of the Navier-Stokes flow around the Block case. Topological changes the loss landscape undergoes as the percentage of sparse data introduced within the training regime is increased. The minima becomes deeper defined and the optimiser arrives at it at much faster as we provide more sparse data to regulate. The X and Y axes represent projection of the trained neural network within orthogonal directions within the weight space, while the Z axis represents the logarithm of the loss value along these projections.}
        \label{fig:landscapes_ns_many}
\end{figure}

The impact of regulation that sparse data performs on the loss landscape can be further elucidated by adding more sparse samples to the PINN. We notice that as we increase the amount of sparse data, the landscape becomes more clearly defined minima and the model arrives at them sooner, hence providing better convergence. Figure \ref{fig:landscapes_ns_many} plots the difference across PINNs regulated with varying amounts of sparse data. 

\section{Coarse Regulated PINNs}
\label{coarse pinns}
In most scenarios, sparse simulation data might not be available or would be rather computationally intensive in obtaining them. It also might seem counter-intuitive to generate sparse simulation data with a numerical simulator before using that in a PINN to converge to the actual solution. However the regulation properties are not limited to sparse representations alone and can be extended to other forms of data that we can introduce within the training regime. \\

A more pragmatic approach would be to use a coarse simulator to generate the training data to be used as a regulator. Coarse simulation data built relatively inexpensively could be used to perform the necessary topological modifications on the loss landscape, allowing the optimiser to converge better and faster. Since PINNs are essentially mesh-invariant the residual loss functions within the training regime would help the PINN generate solutions of arbitrary meshes \cite{LI2018}. This allows for a coupled approach where quick, coarse approximations built by the numerical solver are then fine-tuned with the aid of a PINN. \\

\begin{figure}[h!]
     \centering
     \begin{subfigure}{0.45\textwidth}
         \includegraphics[scale=0.3]{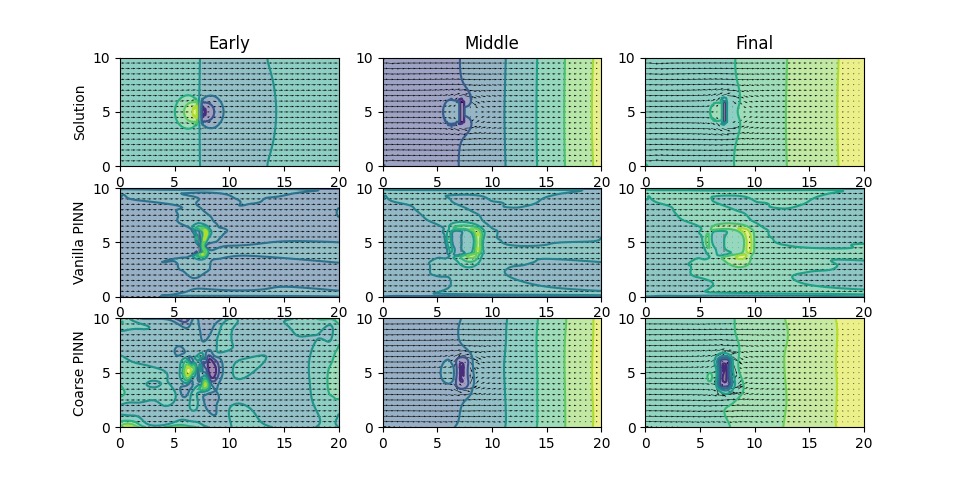}
         \caption{Coarse regulated PINN solution}
         \label{fig:ns_coarse}
     \end{subfigure}
     \hfill
     \begin{subfigure}{0.5\textwidth}
         \includegraphics[scale=0.3]{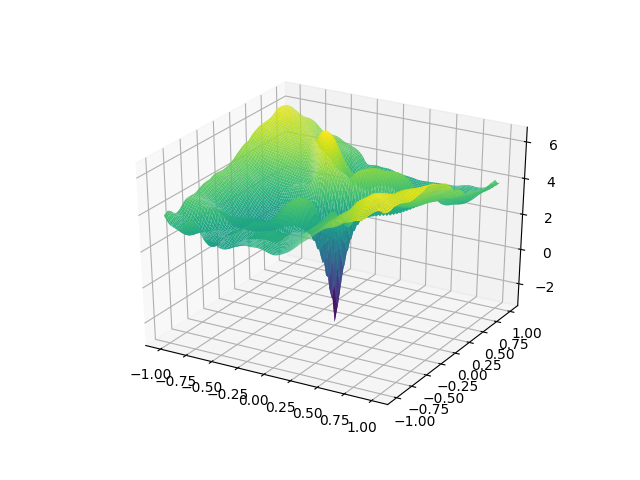}
         \caption{Loss landscape of the coarse PINN}
         \label{fig:ns_coarse_ll}
     \end{subfigure}
        \caption{Figure \ref{fig:ns_coarse} visualises and compares the PINN solutions to the numerical solution. The plot shows the network outputs at the time = 0.01 (early), time=0.5 (middle) and time=1.0 (final). The Vanilla PINN trains to a L2 Error of 8.237 while the Coarse Regulated PINN converges to 0.2135. Figure \ref{fig:ns_coarse_ll} shows the loss landscape of a Coarse regulated PINN. Coarse data is from a mesh with a discretisation 10 times smaller than the numerical solution.}
        \label{fig:landscapes_ns}
\end{figure}

Experimenting on the previous Navier-Stokes equation modelling flow around a block, we notice that the PINN with coarse data introduces similar regulation effects as that with the PINN with sparse data (see figure \ref{fig:ns_coarse}). Data to help with the regulation was built from a coarse solver with a mesh 1/10$^{th}$ coarser than the mesh on which the numerical solution was built and the coarse regulated PINN was tested on. The regulation effects is clearly seen when plotting the loss landscape, with clear defined minima fairly accessible to the minimiser (see figure \ref{fig:ns_coarse_ll} ). 

\section{Experimental Data as the Regulator}
\label{experimental data}
To test further the impact that data regulation can have on the loss landscape, we conducted experiments where we sampled data points from the simulation that is lying along a specific line within the domain. These location specific points mimic linear profiles of data captured by a certain diagnostic device. Our focus remains on the Navier-Stokes flow around the block case. We sampled two linear data points vertically along y-axis at two fixed x positions 1.5 metres before and after the block. From the entirety of the simulation data, we sampled these linear points for every 10$^{th}$ time iteration. We intend for these data points to be treated as linear profiles of experimental data capturing the information of the flow before and after the block.

\begin{figure}[h!]
     \centering
     \begin{subfigure}{0.45\textwidth}
         \includegraphics[scale=0.2]{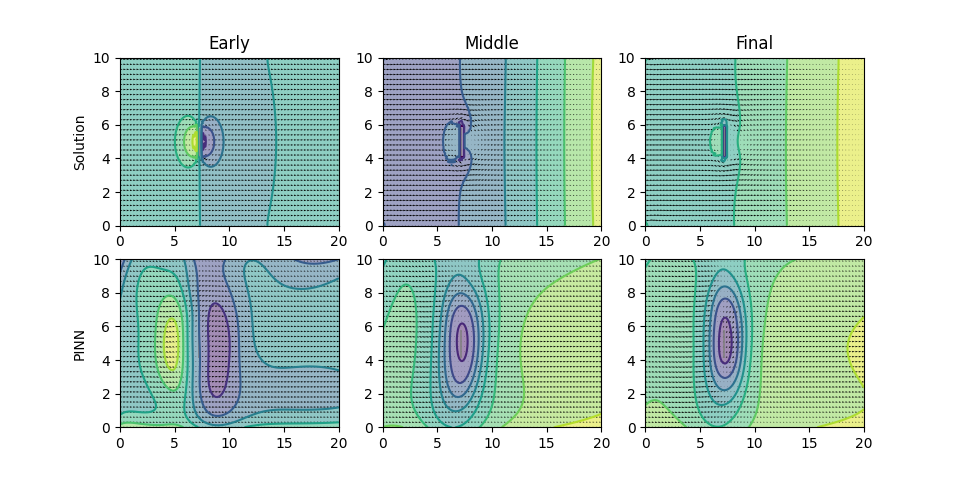}
         \caption{Experimental Data regulated PINN solution}
         \label{fig:ns_line}
     \end{subfigure}
     \hfill
     \begin{subfigure}{0.5\textwidth}
         \includegraphics[scale=0.3]{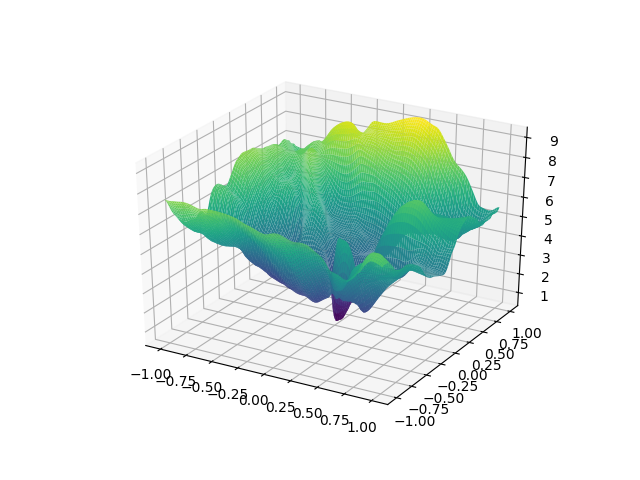}
         \caption{Loss landscape of the Experimental Data regulated PINN}
         \label{fig:ns_line_ll}
     \end{subfigure}
        \caption{Figure \ref{fig:ns_line} visualises and compares the PINN solutions to the numerical solution. The plot shows the network outputs at the time = 0.01 (early), time=0.5 (middle) and time=1.0 (final). The Vanilla PINN trains to a L2 Error of 8.237 while the Data Regulated PINN converges to 1.763. Figure \ref{fig:ns_coarse_ll} shows the loss landscape of the PINN regulated with the linearly captured experimental data.}
        \label{fig:landscapes_ns}
\end{figure}

Though the introduction of the linearly sampled "experimental" data did not help the PINN converge towards the solution, it allowed the network to learn a gross generalisation of the dynamics associated with the losses. As can be seen by the mapping of the loss landscape in figure \ref{fig:ns_line_ll}, the data allows for regulating the landscape to a better degree than to the case without any data. However, the data seems to lack sufficient impact on the loss landscape for the network to converge towards the true solution. \\

\section{Conclusion}
\label{conclusion}

Training a Physics-Informed Neural Network to converge towards the solution of a well-defined PDE is often burdened with several challenges. The network is constrained to operate at a trade-off amongst approximation, generalisation and optimisation errors. By way of introducing regulating data into the training regime, the loss landscape topology is morphed sufficiently to reduce the approximation and optimisation errors. Data regulation for loss landscape engineering allows for taking a hybrid unsupervised-supervised approach that gives sufficient flexibility for the optimisation of the network. Through the course of this paper, we have demonstrated how challenging the loss landscapes of PINNs can be for the minimiser to traverse and how they can be engineered for better performance by way of adding regulating data. \\

We demonstrate that data representative of the PDE solution captured in any form, either sparsely sampled from existing datasets, or built using an inexpensive solver built on a coarse grid across the domain, or even experimental data captured from a specific point within the domain can help modify the landscape. They regulate and ease the optimisation challenge by allowing for the generation of local minima with much better proximity within the hyperspace of the network weights. We demonstrate that utilising the same computational resources, PINNs can improve performance by an order of magnitude without additional increase in training time. Aside from demonstrating methodologies for achieving better convergence in PINNs, we also explore and discuss the potential pitfalls that make training PINNs to the PDE solution a considerable challenge. \\

The relative impact of the data on the objective function should be weighted in its implementation within the training regime. Arguments could be raised to question the need for a PINN if it comes with prerequisite of having solution data. This could be countered by considering that sparse, coarse or even experimental data only produces half truths, partial aspects of the solution that allows to provide the necessary bias for the network. Once the network has been biased towards a neighbourhood much closer to that of the solution within the weight hyperspace, the data regulators can be slowly waned off, allowing for the network to fine-tune to the actual PDE solution. \\

Our next focus of research would be to experiment this method of data-regulated PINNs with experimental data captured from the Tokamak diagnostics in modelling Plasma configurations, exploring a hybrid surrogate model that can marry the physics governed by the PDEs while amalgamating the information found in experimental data. 

% Acknowledgements should go at the end, before appendices and references

\acks{This work has been funded by the EPSRC Energy Programme [grant number  EP/T012250/1]. To obtain further information on the data and models underlying this paper please contact PublicationsManager@ukaea.uk}

% Manual newpage inserted to improve layout of sample file - not
% needed in general before appendices/bibliography.

\newpage

% \appendix
% \section*{Appendix A.}
% \label{app:theorem}

% % Note: in this sample, the section number is hard-coded in. Following
% % proper LaTeX conventions, it should properly be coded as a reference:

% %In this appendix we prove the following theorem from
% %Section~\ref{sec:textree-generalization}:

% In this appendix we prove the following theorem from
% Section~6.2:

% \noindent
% {\bf Theorem} {\it Let $u,v,w$ be discrete variables such that $v, w$ do
% not co-occur with $u$ (i.e., $u\neq0\;\Rightarrow \;v=w=0$ in a given
% dataset $\dataset$). Let $N_{v0},N_{w0}$ be the number of data points for
% which $v=0, w=0$ respectively, and let $I_{uv},I_{uw}$ be the
% respective empirical mutual information values based on the sample
% $\dataset$. Then
% \[
% 	N_{v0} \;>\; N_{w0}\;\;\Rightarrow\;\;I_{uv} \;\leq\;I_{uw}
% \]
% with equality only if $u$ is identically 0.} \hfill\BlackBox

% \noindent
% {\bf Proof}. We use the notation:
% \[
% P_v(i) \;=\;\frac{N_v^i}{N},\;\;\;i \neq 0;\;\;\;
% P_{v0}\;\equiv\;P_v(0)\; = \;1 - \sum_{i\neq 0}P_v(i).
% \]
% These values represent the (empirical) probabilities of $v$
% taking value $i\neq 0$ and 0 respectively.  Entropies will be denoted
% by $H$. We aim to show that $\fracpartial{I_{uv}}{P_{v0}} < 0$....\\

% {\noindent \em Remainder omitted in this sample. See http://www.jmlr.org/papers/ for full paper.}

% \vskip 0.2in
\bibliography{sample}

\begin{thebibliography}{15}
\providecommand{\natexlab}[1]{#1}
\providecommand{\url}[1]{\texttt{#1}}
\expandafter\ifx\csname urlstyle\endcsname\relax
  \providecommand{\doi}[1]{doi: #1}\else
  \providecommand{\doi}{doi: \begingroup \urlstyle{rm}\Url}\fi

\bibitem[Andasari(2020)]{wave_spectral}
Vivi Andasari.
\newblock {Numerical Methods using Python}.
\newblock
  \url{http://people.bu.edu/andasari/courses/numericalpython/python.html},
  2020.
\newblock [Online; accessed 9-July-2021].

\bibitem[Baydin et~al.(2018)Baydin, Pearlmutter, Radul, and
  Siskind]{BAYDIN2018}
Atilim~Gunes Baydin, Barak~A Pearlmutter, Alexey~Andreyevich Radul, and
  Jeffrey~Mark Siskind.
\newblock Automatic differentiation in machine learning: a survey.
\newblock \emph{Journal of machine learning research}, 18, 2018.

\bibitem[Blum and Rivest(1992)]{BLUM1992}
Avrim~L. Blum and Ronald~L. Rivest.
\newblock Training a 3-node neural network is np-complete.
\newblock \emph{Neural Networks}, 5\penalty0 (1):\penalty0 117--127, 1992.
\newblock ISSN 0893-6080.
\newblock \doi{https://doi.org/10.1016/S0893-6080(05)80010-3}.
\newblock URL
  \url{https://www.sciencedirect.com/science/article/pii/S0893608005800103}.

\bibitem[Bottou and Bousquet(2008)]{BOTTOU2008}
Léon Bottou and Olivier Bousquet.
\newblock The tradeoffs of large scale learning.
\newblock In \emph{Advances in Neural Information Processing Systems}, pages
  161--168, 2008.
\newblock URL \url{http://leon.bottou.org/publications/pdf/nips-2007.pdf}.

\bibitem[Goldstein and Studer(2016)]{GOLDSTEIN2016}
Tom Goldstein and Christoph Studer.
\newblock Phasemax: Convex phase retrieval via basis pursuit, 2016.
\newblock URL \url{https://arxiv.org/abs/1610.07531}.

\bibitem[He et~al.(2015)He, Zhang, Ren, and Sun]{HE2015}
Kaiming He, Xiangyu Zhang, Shaoqing Ren, and Jian Sun.
\newblock Deep residual learning for image recognition.
\newblock \emph{CoRR}, abs/1512.03385, 2015.
\newblock URL \url{http://arxiv.org/abs/1512.03385}.

\bibitem[Hennigh et~al.(2020)Hennigh, Narasimhan, Nabian, Subramaniam,
  Tangsali, Rietmann, del Aguila~Ferrandis, Byeon, Fang, and
  Choudhry]{HENNIGH2020}
Oliver Hennigh, Susheela Narasimhan, Mohammad~Amin Nabian, Akshay Subramaniam,
  Kaustubh Tangsali, Max Rietmann, Jose del Aguila~Ferrandis, Wonmin Byeon,
  Zhiwei Fang, and Sanjay Choudhry.
\newblock Nvidia simnet: an ai-accelerated multi-physics simulation framework,
  2020.

\bibitem[Kingma and Ba(2017)]{KINGMA2017}
Diederik~P. Kingma and Jimmy Ba.
\newblock Adam: A method for stochastic optimization, 2017.

\bibitem[Li et~al.(2018)Li, Xu, Taylor, Studer, and Goldstein]{LI2018}
Hao Li, Zheng Xu, Gavin Taylor, Christoph Studer, and Tom Goldstein.
\newblock Visualizing the loss landscape of neural nets, 2018.

\bibitem[Lu et~al.(2019)Lu, Meng, Mao, and Karniadakis]{LU2019}
Lu~Lu, Xuhui Meng, Zhiping Mao, and George~E. Karniadakis.
\newblock Deepxde: {A} deep learning library for solving differential
  equations.
\newblock \emph{CoRR}, abs/1907.04502, 2019.
\newblock URL \url{http://arxiv.org/abs/1907.04502}.

\bibitem[Ngatchou et~al.(2005)Ngatchou, Zarei, and El-Sharkawi]{NGATCHOU2005}
P.~Ngatchou, A.~Zarei, and A.~El-Sharkawi.
\newblock Pareto multi objective optimization.
\newblock In \emph{Proceedings of the 13th International Conference on,
  Intelligent Systems Application to Power Systems}, pages 84--91, 2005.
\newblock \doi{10.1109/ISAP.2005.1599245}.

\bibitem[Raissi et~al.(2019)Raissi, Perdikaris, and Karniadakis]{RAISSI2019}
M.~Raissi, P.~Perdikaris, and G.E. Karniadakis.
\newblock Physics-informed neural networks: A deep learning framework for
  solving forward and inverse problems involving nonlinear partial differential
  equations.
\newblock \emph{Journal of Computational Physics}, 378:\penalty0 686--707,
  2019.
\newblock ISSN 0021-9991.
\newblock \doi{https://doi.org/10.1016/j.jcp.2018.10.045}.
\newblock URL
  \url{https://www.sciencedirect.com/science/article/pii/S0021999118307125}.

\bibitem[Ramachandran et~al.(2017)Ramachandran, Zoph, and Le]{RAMACHANDRAN2017}
Prajit Ramachandran, Barret Zoph, and Quoc~V. Le.
\newblock Searching for activation functions.
\newblock \emph{CoRR}, abs/1710.05941, 2017.
\newblock URL \url{http://arxiv.org/abs/1710.05941}.

\bibitem[Sun et~al.(2020)Sun, Gao, Pan, and Wang]{SUN2020}
Luning Sun, Han Gao, Shaowu Pan, and Jian-Xun Wang.
\newblock Surrogate modeling for fluid flows based on physics-constrained deep
  learning without simulation data.
\newblock \emph{Computer Methods in Applied Mechanics and Engineering},
  361:\penalty0 112732, 2020.
\newblock ISSN 0045-7825.
\newblock \doi{https://doi.org/10.1016/j.cma.2019.112732}.
\newblock URL
  \url{https://www.sciencedirect.com/science/article/pii/S004578251930622X}.

\bibitem[Tannehill(1997)]{TANNEHILL1997}
John~C Tannehill.
\newblock \emph{Computational Fluid Mechanics and Heat Transfer}.
\newblock Taylor and Francis, Washington D.C., 1997.

\end{thebibliography}

\end{document}